\documentclass{article}

\usepackage[preprint]{neurips_2021}

\usepackage[utf8]{inputenc} 
\usepackage[T1]{fontenc}    
\usepackage{hyperref}       
\usepackage{url}            
\usepackage{booktabs}       
\usepackage{amsfonts}       
\usepackage{nicefrac}       
\usepackage{microtype}      
\usepackage{xcolor}         

\usepackage{amsmath}
\usepackage{algorithm}
\usepackage{algorithmic}
\usepackage{makecell}
\usepackage{mathtools}
\usepackage{threeparttable}
\usepackage{amssymb}
\usepackage{amsthm}
\newtheorem{example}{Example}
\newtheorem{theorem}{Theorem}
\newtheorem{definition}{Definition}
\newtheorem{lemma}{Lemma}
\newtheorem{remark}{Remark}

\title{Towards Sharper Utility Bounds for Differentially Private Pairwise Learning}

\author{%
  Yilin Kang \\
  Institute of Information Engineering, Chinese Academy of Sciences \\
  School of Cyber Security, University of Chinese Academy of Sciences \\
  \texttt{kangyilin@iie.ac.cn} \\
  \And
  Yong Liu \thanks{corresponding author} \\
  Renmin University of China \\
  \texttt{liuyonggsai@ruc.edu.cn} \\
  \AND
  Jian Li \\
  Institute of Information Engineering, Chinese Academy of Sciences \\
  \texttt{lijian9026@iie.ac.cn} \\
  \And
  WeiPing Wang \\
  Institute of Information Engineering, Chinese Academy of Sciences \\
  \texttt{wangweiping@iie.ac.cn} \\
}

\begin{document}

\maketitle

\begin{abstract}
  Pairwise learning focuses on learning tasks with pairwise loss functions, depends on pairs of training instances, and naturally fits for modeling relationships between pairs of samples.
  In this paper, we focus on the privacy of pairwise learning and propose a new differential privacy paradigm for pairwise learning, based on gradient perturbation.
  Except for the privacy guarantees, we also analyze the excess population risk and give corresponding bounds under both expectation and high probability conditions.
  We use the \textit{on-average stability} and the \textit{pairwise locally elastic stability} theories to analyze the expectation bound and the high probability bound, respectively.
  Moreover, our analyzed utility bounds do not require convex pairwise loss functions, which means that our method is general to both convex and non-convex conditions.
  Under these circumstances, the utility bounds are similar to (or better than) previous bounds under convexity or strongly convexity assumption, which are attractive results.
\end{abstract}

\section{Introduction}

Pointwise learning and pairwise learning are two basic learning problems in the field of Empirical Risk Minimization (ERM).
Traditional pointwise learning takes individual samples as the input and pairwise learning studies the condition that taking pairs of samples as input.
In the scenarios such as classification, regression, pointwise learning is popular and is shown effective (\cite{5}).
And pairwise learning naturally fits the situation that the connection between two samples matters (\cite{1}), such as in the fields of clustering (\cite{2}), metric learning (\cite{6}), ranking (\cite{3}), etc.

Although both pointwise and pairwise learning frameworks have obtained great success in machine learning, they are facing huge privacy problems because training a single machine learning model needs large quantities of personal data.
Actually, not only the leakage of original data will disclose the information of individuals, when training machine learning models, model parameters may indirectly reveal sensitive information as well (\cite{8,9}).
Under these circumstances, Differential Privacy (DP) (\cite{7}) is proposed as a strong mathematical scheme for privacy preservation, especially in machine learning algorithms.
As a result, it has been widely studied in recent decades.
Differential privacy preserves sensitive information by adding random noise (called perturbation) when training machine learning models, to make the adversaries cannot infer any data instance owned by a certain individual after getting model parameters.

There is a long list of works on Differentially Private Empirical Risk Minimization (DP-ERM) of pointwise learning (\cite{10,11,15,12,16,13,14}).
In traditional settings, three perturbation methods were proposed: output perturbation, objective perturbation, and gradient perturbation.
However, for pairwise learning, the studies on differential privacy is rare.
\citet{17} discussed the differential privacy of pairwise learning and analyzed the utility under both online and offline settings.
For offline setting, it directly added the random noise to the final model and can be seemed as an output perturbation method.
\cite{4} proposed an output perturbation method for DP pairwise learning under non-smooth condition, matching the results given in \cite{17} under smooth condition.
Moreover, \citet{17,4} assumed that the loss function to be convex (or even strongly convex), which is not easy to hold in some situations.

In this paper, we propose a gradient perturbation method for DP pairwise learning, the advantages include:
1) Gradient descent is a widely used optimization method, thus the gradient perturbation method can be used for a wide range of applications;
2) The gradient perturbation method not only protects the model, but also the gradient;
3) The gradient perturbation method allows the model to escape local minima (\cite{21}).
The main contributions of this paper include:\\
\textbf{A Gradient Perturbation Method.}
We propose a gradient perturbation differential privacy method for pairwise learning and give the privacy guarantees.
To the best of our knowledge, this is the first `gradient-based' method in the field of differentially private pairwise learning.
Moreover, the utility analysis of gradient perturbation method is more complicated than output one, because random noise is added $T$ times in gradient pertubation method and only one time directly at the final model in output perturbation methods.\\
\textbf{Pairwise Locally Elastic Stability.}
To get the high probability excess population risk bound, we first propose the \textit{pairwise influence function} and find it shares the same property as the traditional pointwise influence function (\cite{29}).
Then, via pairwise influence function, we extend the locally elastic stability (\cite{20}) to pairwise learning condition, called \textit{pairwise locally elastic stability}, to analyze the high probability utility bound. \\
\textbf{Non-convex Utility Analysis.}
We theoretically analyze the excess population risk and give corresponding bounds under both expectation and high probability conditions.
For expectation utility bound, we introduce the \textit{on-average stability} (\cite{24}) to the field of differential privacy and get an excess population risk bound of $\mathcal{O}\left(\sqrt{p}/(n\epsilon)\right)$.
For the high probability bound, we propose \textit{pairwise locally elastic stability} and get an $\mathcal{O}\left(\sqrt{p}/(\sqrt{n}\epsilon)\right)$ excess population risk bound.
Additionally, the utility bounds do not assume the loss function to be convex.
Under these circumstances, detailed theoretical analysis shows that we achieve similar or better theoretical results, which is attractive.

The rest of the paper is organized as follows.
First, we introduce some previous work in Section 2.
Then, some preliminaries are presented in Section 3.
We propose the gradient perturbation pairwise learning algorithm in detail and give the privacy guarantees in Section 4.
We theoretically analyze the utility bounds in Section 5.
Detailed comparisons with previous works are shown in Section 6.
Some examples are given in Section 7.
Finally, we conclude the paper in Section 8.

\section{Related Work}

In the last decade, there is a long list of papers focusing on DP pointwise learning problems: \citet{10,11,12} are for convex conditions and in \cite{13,14,16}, non-convex conditions are taken into account.
The approaches of adding noise include the output perturbation method, the objective perturbation method, and the gradient perturbation method.
However, all of the above methods are designed for pointwise loss functions.

The independently identically distribution (i.i.d) assumption is a basic assumption in pointwise learning.
However, it is not applicable for pairwise learning, so existed methods mentioned above can not be easily extended to pairwise learning.
Under these circumstances, \citet{17} considered output perturbation method for both online and offline settings in the field of pairwise learning.
\citet{4} extended the output perturbation method to the condition that the loss function is not assumed to be smooth.
However, in \cite{17,4}, only convex pairwise loss functions are studied and the results cannot be easily extended to non-convex conditions.
Since gradient descent is a widely used optimization method for machine learning algorithms, we propose a gradient perturbation method for pairwise learning in this paper.
Meanwhile, we analyze the excess population risk and give corresponding bounds under both expectation and high probability conditions.
Moreover, our given utility bounds do not assume the loss function to be convex, which are more general results.
Theoretical analysis shows that the utility bounds of our method are similar to (or even better than) which proposed by \cite{17,4} under convex and strongly convex conditions.

\section{Preliminaries}

\subsection{Differential Privacy}

Databases $D,D'\in\mathcal{D}^n$ differing by one single entry are denoted as $D\sim D'$, called \textit{adjacent databases}.
For a given vector $x=[x_1,...,x_d]^T$, its $\ell_2$-norm is: $\left\Vert x\right\Vert_2=(\sum_{i=1}^{d}|x_i|^2)^{\frac{1}{2}}$.

\begin{definition}[Differential Privacy (\cite{7})]
A randomized algorithm $\mathcal{A}:\mathcal{D}^n\rightarrow\mathbb{R}^p$ is ($\epsilon,\delta$)-differential privacy (($\epsilon,\delta$)-DP) if for all $D\sim D'$ and all events $S\in range(\mathcal{A})$:
\begin{equation*}
\mathbb{P}\left[\mathcal{A}(D)\in S\right]\leq e^\epsilon\mathbb{P}\left[\mathcal{A}(D')\in S\right]+\delta.
\end{equation*}
\end{definition}

Differential privacy requires adjacent datasets leading to similar distributions on the output of algorithm $\mathcal{A}$.
This implies that adversaries cannot infer whether an individual participates in the training process because essentially the same conclusions about a certain individual will be drawn whether or not that individual’s data was used.

\subsection{Pairwise Learning}

In traditional pointwise learning problems, the loss function is defined as $\ell:\mathcal{C}\times\mathcal{D}\rightarrow\mathbb{R}$, where $\mathcal{C}\subseteq\mathbb{R}^p$ is a parameter space of dimension $p$ and  $\mathcal{D}=\mathcal{X}\times\mathcal{Y}$ is the data universe, $\mathcal{X}\subset\mathbb{R}^d$ is an input space of dimension $d$ and $\mathcal{Y}\subset\mathbb{R}$ is an output space.
Pairwise loss functions on pairs of data records are: $\ell:\mathcal{C}\times\mathcal{D}\times\mathcal{D}\rightarrow\mathbb{R}$.
Given a training dataset $D=\{z_1,...,z_n\}\subseteq\mathcal{D}^n$ where $z_i=(x_i, y_i)$ and a loss function $\ell(\cdot;\cdot,\cdot)$, with $\theta\in\mathcal{C}$, the pairwise empirical risk is:
\begin{equation}\label{pairwiseobj}
L(\theta;D)=\frac{1}{n(n-1)}\sum_{i=1}^{n}\sum_{j\neq i}\ell(\theta;z_i,z_j).
\end{equation}
Assuming that the inputs are drawn from an underlying distribution $\mathcal{P}$, the population risk is:
\begin{equation*}
L_\mathcal{P}(\theta)=\mathbb{E}_{z_i,z_j\sim\mathcal{P}}\left[\ell(\theta;z_i,z_j)\right].
\end{equation*}

Private pairwise learning is to find a private $\theta_{priv}$ making the algorithm ($\epsilon,\delta$)-DP, and the error is minimized.
In this paper, we aim to minimize the population risk mentioned above.
The performance of $\theta_{priv}$ is quantified by the excess population risk:
\begin{align}\label{EPR}
Err_\mathcal{P}(\theta_{priv})=L_\mathcal{P}(\theta_{priv})-\min_{\theta\in\mathcal{C}}L_\mathcal{P}(\theta).
\end{align}
For simplicity, we denote $\theta_\mathcal{P}^*=\arg\min_{\theta\in\mathcal{C}}L_\mathcal{P}(\theta)$ and $\theta^*=\arg\min_{\theta\in\mathcal{C}}L(\theta;D)$.

Thus, (\ref{EPR}) can be decomposed as:
\begin{equation}\label{decomEPR}
Err_\mathcal{P}(\theta_{priv})=\left[L_\mathcal{P}(\theta_{priv})-L\left(\theta_{priv};D\right)\right]+\left[L(\theta_{priv};D)-L(\theta^*;D)\right]+\left[L(\theta^*;D)-L_\mathcal{P}(\theta_\mathcal{P}^*)\right].
\end{equation}
The first term on the right hand side of (\ref{decomEPR}) is the generalizetion error.
In this paper, we use algorithmic stability theory to bound it.
The second term is called excess empirical risk, we use optimization theory to solve it.

\section{Differentially Private Pairwise Learning by Gradient Perturbation}

The gradient descent method is widely used in the field of machine learning in recent decades.
As a result, the gradient perturbation method is feasible for most algorithms.
Moreover, by applying the gradient perturbation method, both the final model and the gradients can be protected, which is more reliable for privacy.
Besides, gradient perturbation method can allow the model to escape local minima (\cite{21}).
Thus, we focus on the gradient perturbation method in this paper.

When applying gradient descent method, the parameter update process at the $t^{th}$ iteration is:
\begin{equation*}
\theta_{t}=\theta_{t-1}-\eta_t\nabla_\theta L(\theta_{t-1};D),
\end{equation*}
where $L(\theta;D)$ is the same as in (\ref{pairwiseobj}).

\begin{definition}[Gradient perturbation in pairwise learning]
When it comes to gradient perturbation method, the gradient descent process at iteration $t$ becomes to:
\begin{equation*}
\theta_{t}=\theta_{t-1}-\eta_t\left(\nabla_\theta L(\theta_{t-1};D)+b_t\right),
\end{equation*}
where $b_t$ denotes the random noise to preserve the privacy at iteration $t$.
\end{definition}

Our proposed gradient perturbation DP-ERM method for pairwise learning is detailed in Algorithm 1, in which the pairwise loss function is assumed $G$-Lipschitz, defined as:
\begin{definition}[$G$-Lipschitz]
	A pairwise loss function $\ell:\mathcal{C}\times\mathcal{D}\times\mathcal{D}\rightarrow\mathbb{R}$ is $G$-Lipschitz over $\theta$ if for any $z,z'\in\mathcal{D}$ and $\theta_1,\theta_2\in\mathcal{C}$, the following inequality holds:
	\begin{equation*}
	|\ell(\theta_1;z,z')-\ell(\theta_2;z,z')|\leq G\left\Vert\theta_1-\theta_2\right\Vert_2.
	\end{equation*}
\end{definition}

Meanwhile, in Algorithm 1, each dimension of $b_t$ is chosen independently at each iteration $t$.

\begin{algorithm}[tb]
\caption{Gradient perturbation pairwise learning}
\label{alg1}
\textbf{Input}: Privacy budgets $\epsilon,\delta$, dataset $D=\{z_1,...,z_n\}$, constrained set $\mathcal{C}\subset\mathbb{R}^p$, pairwise loss function $\ell(\cdot;\cdot,\cdot)$, total number of iterations $T$ and learning rate $\eta_t$ for each iteration.\\
\textbf{Parameter}: Loss function $\ell(\cdot;\cdot,\cdot)$ is $G$-Lipschitz over $\theta$.\\
\textbf{Output}: Differentially private model $\theta_{priv}$.

\begin{algorithmic}[1]
\STATE Let $t=1$, randomly set $\theta_0$.
\WHILE{$t\leq T$}
\STATE $\theta_t\leftarrow\theta_{t-1}-\eta_t\left(\frac{1}{n(n-1)}\sum_{i=1}^{n}\sum_{j\neq i}\nabla_\theta\ell(\theta_{t-1};z_i,z_j)+b_t\right)$, where $b_t\sim\mathcal{N}\left(0,\sigma_t^2I_p\right)$.
\ENDWHILE
\STATE $\theta_{priv}=\theta_T$.
\STATE \textbf{return} $\theta_{priv}$
\end{algorithmic}
\end{algorithm}

\subsection{Privacy Guarantees}

\begin{theorem}
Algorithm 1 satisfies ($\epsilon,\delta$)-DP if the pairwise loss fucntion is $G$-Lipschitz over $\theta$ and:
\begin{equation*}
\sigma_t\geq\frac{8G\sqrt{T\log(1/\delta)}}{n\epsilon}.
\end{equation*}
\end{theorem}

Detailed proof is given in Appendix B.1.
We analyze the privacy via the moments accountant method (\cite{18}), the difference is that we extend it to the pairwise setting.
For completeness, we also analyze the privacy guarantees from another point of view: applying the fundamental Gaussian mechanism to the $\ell_2$-sensitivity of the gradient.
Details are shown in Appendix B.2.
Fundamental Gaussian mechanism gives a noise bound (represented by the variance) of $\mathcal{O}\big(\frac{T^2\log(T/\delta)}{n^2\epsilon^2}\big)$.
It is $\mathcal{O}\big(\frac{T\log(T/\delta)}{\log(1/\delta)}\big)$ times looser than the result discussed by the moments accountant method.

\begin{remark}
\rm{
To guarantee privacy, we only assume that $\ell\left(\cdot;\cdot,\cdot\right)$ is $G$-Lipschitz.
However, in \cite{17}, except for $G$-Lipschitz, $L$-smooth and (strongly) convex are required when discussing the privacy.
}
\end{remark}

In Theorem 1, $\sigma_t$ is the same for each iteration $t$, so we use $\sigma$ as a shorthand for $\sigma_t$ in the following.

\section{Utility Bounds}

In this section, we theoretically analyze the \textbf{expectation} and \textbf{high probability} utility bounds.
As shown in (\ref{decomEPR}), the excess population risk is decomposed to the excess empirical risk and the generalization error, so we also analyze them, details are shown in the Appendix.

We now introduce some necessary assumptions, which will be applied throughout the following.

\begin{definition}[$L$-smooth]
	A pairwise loss function $\ell:\mathcal{C}\times\mathcal{D}\times\mathcal{D}\rightarrow\mathbb{R}$ is $L$-smooth over $\theta$ if for any $z,z'\in\mathcal{D}$ and $\theta_1,\theta_2\in\mathcal{C}$, the following inequality holds:
	\begin{equation*}
	\|\nabla_\theta\ell(\theta_1;z,z')-\nabla_\theta\ell(\theta_2;z,z')\|_2\leq L\left\Vert\theta_1-\theta_2\right\Vert_2.
	\end{equation*}
\end{definition}

\begin{definition}[Polyak-{\L}ojasiewicz condition]
	Pairwise objective function $L(\theta;D)$ satisfies the Polyak-{\L}ojasiewicz (PL) condition with parameter $\mu$, if there exists $\mu>0$ for all $\theta$:
	\begin{equation*}
	\left\Vert\nabla_\theta L(\theta;D)\right\Vert_2^2\geq 2\mu\left(L(\theta;D)-\min_{\theta\in\mathcal{C}}L(\theta;D)\right).
	\end{equation*}
\end{definition}

If the pairwise objective function is strongly convex, then it naturally satisfies PL inequality.
However, comparing with \textit{(strongly) convex}, PL condition is more general (\cite{32}) and can be guaranteed under some non-convex conditions.
There are some examples of functions satisfying PL condition, including neural networks with one hidden layer, ResNets, and objective functions in matrix factorization (\cite{28}).
In this paper, we mainly use the Polyak-{\L}ojasiewicz inequality to release the convexity assumption.

\subsection{Expectation Bound}
\begin{theorem}
	Assuming that the pairwise loss function is $G$-Lipschitz, $L$-smooth over $\theta$, and the pairwise empirical risk satisfies the PL condition with parameter $\mu$. For all iterations $t$, the learning rates are the same, i.e. $\eta_1=...=\eta_T=\eta=\frac{1}{L}$, then the excess population risk satisfies:
	\begin{equation*}
	\begin{aligned}
	\mathbb{E}\left[L_\mathcal{P}(\theta_{priv})-\min_{\theta\in\mathcal{C}}L_\mathcal{P}(\theta)\right]&\leq\inf_{\tau>0}\left\{2(\tau+L)\left(\frac{16G^2T^2}{L^2n^2}+\frac{pT^2\sigma^2}{2L^2}+\frac{4\sqrt{2}\sqrt{p}GT^2\sigma}{L^2n}\right)\right. \\
	&~~~~~~~~~~~~~~\left.+\frac{G^2}{2\tau}+\left(1-\frac{\mu}{L}\right)^T\left(L(\theta_0;D)-L(\theta^*;D)\right)+\frac{p\sigma^2}{2\mu}\right\}.
	\end{aligned}
	\end{equation*}
\end{theorem}

Detailed proof can be found in Appendix C.1.
The proof is motivated by \cite{24} in the strongly convex non-private case but more involved in differential privacy and non-convex case.
The key challenges include that without convexity assumption, the gap between model parameters derived from adjacent datasets: $\|\theta^*-\theta_i^*\|_2$ cannot be easily bounded.
And in differential privacy, especially for gradient perturbation method, the model parameters are perturbed more than once, which are not easy to measure.

\begin{remark}
	\rm{
	By the definitions of $\mu$ and $L$, it is easy to follow that $0<\mu<L$ (\cite{13}).
	Meanwhile, with $0\leq\ell\left(\cdot;\cdot,\cdot\right)\leq M_\ell$, $L(\theta_0;D)-L(\theta^*;D)$ is bounded.
	As a result, term $\left(1-\frac{\mu}{L}\right)^T\left(L(\theta_0;D)-L(\theta^*;D)\right)$ on the right side of inequality becomes smaller with the increasing of $T$.
	Note that the inequality holds for all $\tau>0$, so if we take $\tau=\mathcal{O}\left(\frac{n\epsilon}{\sqrt{p}T^{1.5}}\right)$, $\sigma=\mathcal{O}\left(\frac{\sqrt{T}}{n\epsilon}\right)$ given by Theorem 1, and $T=\mathcal{O}\left(\log\left(\frac{n\epsilon}{\sqrt{p}}\right)\right)$, the expectation excess population risk bound comes to:
	\begin{equation*}
	\mathcal{O}\left(\frac{\sqrt{p}}{n\epsilon}+\frac{\epsilon}{n\sqrt{p}}+\frac{1}{n}+\frac{p}{n^2\epsilon^2}\right)=\mathcal{O}\left(\frac{\sqrt{p}}{n\epsilon}\right).
	\end{equation*}
}
\end{remark}

To the best of knowledge, this is the first excess population risk bound under non-convex analysis for differentially private pairwise learning, achieving an $\mathcal{O}\left(\frac{\sqrt{p}}{n\epsilon}\right)$ expectation bound.

\subsection{High Probability Bound}
To get the high probability bound, we first propose the pairwise locally elastic stability.
Motivated by \cite{20}, we extend the locally elastic stability to pairwise learning as the following:
\begin{definition}[Pairwise Locally Elastic Stability]
	Algorithm $\mathcal{A}$ has pairwise locally elastic stabiliy $\beta_n\left(\cdot;\cdot,\cdot\right)$ with respect to the pairwise loss function $\ell$ if for all $n$, inequality
	\begin{equation*}
	\left|\ell\left(\mathcal{A}_D;z_j,z_k\right)-\ell\left(\mathcal{A}_{D^{-i}};z_j,z_k\right)\right|\leq\beta_n\left(z_i;z_j,z_k\right)
	\end{equation*}
	holds for all $D\in\mathcal{D}^n,1\leq i\leq n$, and $z_j,z_k\in\mathcal{D}$, where $D^{-i}=\{z_1,\cdots,z_{i-1},z_{i+1},\cdots,z_n\}$ and $\mathcal{A}_D$ denotes the model derived from dataset $D$ by applying $\mathcal{A}$.
\end{definition}

In words, pairwise locally elastic stability gives a certain stability coefficient for each data pair $(z_j,z_k)$, to reflect the change in the pairwise loss due to the removal of $z_i$.
The stability coefficient depends on $z_i,z_j$ and $z_k$.
In other words, pairwise locally elastic stability is data-dependent, which is different from the uniform stability.
To further appreciate this definition, we compare it with pairwise uniform stability.

\begin{definition}[Pairwise Uniform Stability (\cite{27})]
	Algorithm $\mathcal{A}$ has pairwise uniform stabiliy $\beta_n^U$ with respect to the pairwise loss function $\ell$ if
	\begin{equation*}
	\left|\ell\left(\mathcal{A}_D;z,z'\right)-\ell\left(\mathcal{A}_{D^{-i}};z,z'\right)\right|\leq\beta_n^U
	\end{equation*}
	holds for all $D\in\mathcal{D}^n,1\leq i\leq n$, and $z,z'\in\mathcal{D}$.
\end{definition}

By defintion, one can set $\beta_n^U=\sup_{z,z',z''}\beta_n\left(z;z',z''\right)$.
It is easy to follow that the pairwise locally elastic stability offers a finer-grained delineation of the loss function sensitivity.
When the worst $\left|\ell\left(\mathcal{A}_D;z,z'\right)-\ell\left(\mathcal{A}_{D^{-i}};z,z'\right)\right|$ is much larger than the typical realizations, $\beta_n^U$ will be much larger than $\mathbb{E}_{z,z',z''}\beta_n(z;z,z'')$.
Thus, in this case, the pairwise locally elastic stability is much better than the pairwise uniform stability.

\paragraph{Estimation Using Pairwise Influence Functions.}

To estimate pairwise locally elastic stability $\beta_n(z,z',z'')$, we propose \textbf{pairwise influence function}, motivated by \cite{29} in the pointwise setting but more involved in pairwise case.

For pairwise loss function $\ell(\cdot;\cdot,\cdot)$, defining $\mathcal{A}_D^*=\arg\min_{\theta\in\mathcal{C}}\frac{1}{n(n-1)}\sum_{j=1}^{n}\sum_{k\neq j}\ell\left(\theta;z_j,z_k\right)$ and $\mathcal{A}_{D^{-i}}^*=\arg\min_{\theta\in\mathcal{C}}\frac{1}{n(n-1)}\sum_{j=1,j\neq i}^{n}\sum_{k\neq i,j}\ell\left(\theta;z_j,z_k\right)$.
Then, the pairwise influence function of data instance $z_i$ is:
\begin{equation}\label{pairwiseIF}
\mathcal{A}_{D^{-i}}^*-\mathcal{A}_D^*\approx\frac{1}{n(n-1)}\left[\nabla^2_\theta L\left(\mathcal{A}_D^*;D\right)\right]^{-1}\sum_{z'\in D;z'\neq z_i}\left(\nabla_\theta\ell\left(\mathcal{A}_D^*;z_i,z'\right)+\nabla_\theta\ell\left(\mathcal{A}_D^*;z',z_i\right)\right).
\end{equation}

Detailed discussion is shown in Appendix C.2, the key is Taylor expansion (\cite{34}).
By the definitions of $\mathcal{A}_D^*$ and $\mathcal{A}_{D^{-i}}^*$, it can be easily observed that the pairwise influence function reflects the change on the model parameter, due to the removal of a data instance from the dataset.

Now, with pairwise influence function, we are ready to estimate pairwise elastic stability $\beta_n(z;z',z'')$.

\begin{lemma}
Defining $\beta_n\left(z_i;z_j,z_k\right)\coloneqq|\ell\left(\mathcal{A}_{D}^*;z_j,z_k\right)-\ell\left(\mathcal{A}_{D^{-i}}^*;z_j,z_k\right)|$, it is an $\mathcal{O}(1/n)$ term.
\end{lemma}

Via Taylor expansion, we have:
\begin{equation*}
\left|\ell\left(\mathcal{A}_{D}^*;z_j,z_k\right)-\ell\left(\mathcal{A}_{D^{-i}}^*;z_j,z_k\right)\right|\approx\left|\left(\mathcal{A}_{D^{-i}}^*-\mathcal{A}_{D}^*\right)^T\nabla_\theta\ell\left(\mathcal{A}_{D}^*;z_j,z_k\right)\right|.
\end{equation*}

Note that $\ell(\cdot;\cdot,\cdot)$ is $G$-Lipschitz, then $\nabla_\theta\ell(\cdot;\cdot,\cdot)$is bounded, so for many problems one has $\mathcal{A}_{D^{-i}}^*-\mathcal{A}_D^*$ of the order $\mathcal{O}\left(1/n\right)$.
Thus, $|\ell\left(\mathcal{A}_{D}^*;z_j,z_k\right)-\ell\left(\mathcal{A}_{D^{-i}}^*;z_j,z_k\right)|$ is of the order $\mathcal{O}\left(1/n\right)$.
Moreover, \cite{27} shows that $\beta_n^U$ in uniform stability satisfies $\beta_n^U=\mathcal{O}\left(1/n\right)$ for many pairwise problems including kernel-based ranking algorithm with least squares ranking loss and hinge ranking loss.
This fact means that it is reasonable to assume $\beta_n\left(z;z',z''\right)=\mathcal{O}\left(1/n\right)$ for pairwise locally elastic stability.
Then Lemma 1 holds.

More specifically, for some function $\beta\left(\cdot;\cdot,\cdot\right)$ that is independent of $n$, we assume:
\begin{equation*}
\beta_n\left(z_i;z_j,z_k\right)=\frac{\beta\left(z_i;z_j,z_k\right)}{n},
\end{equation*}
which means $\sup_nn\beta_n\left(z_i;z_j,z_k\right)$ is finite for all $z_i,z_j,z_k$.
Last, like discussed in Remark 2, we assume the loss function $\ell$ satisfies $0\leq\ell\leq M_\ell$ for a constant $M_\ell$.

Now, we are ready to state our high probability excess population risk bound.
\begin{theorem}
	If the pairwise loss fucntion is $G$-Lipschitz, $L$-smooth, and the empirical risk satisfies the PL condition with $\mu$ over $\theta$, taking $T=\mathcal{O}\left(\log(\sqrt{n}\epsilon)\right)$, and for all iterations $t$, the learning rates are the same, i.e. $\eta_1=...=\eta_T=\frac{1}{L}=\eta$.
	Then, with probability at least $1-\xi$, we have:
	\begin{equation}\label{the2}
	\begin{aligned}
	&L_\mathcal{P}(\theta_{priv})-\min_{\theta\in\mathcal{C}}L_\mathcal{P}(\theta) \\
	&\leq\frac{2\sup_{z\in\mathcal{D}}\mathbb{E}_{z',z''}\beta\left(z;z',z''\right)}{n}+\left(4\sup_{z\in\mathcal{D}}\mathbb{E}_{z',z''}\left[\beta\left(z;z',z''\right)\right]+4M_\ell\right)\sqrt{\frac{2\log\left(12/\xi\right)}{n}} \\
	&\quad+c_1\frac{G^2}{L\epsilon}\sqrt{\frac{p\log(12/\xi)\log(1/\delta)}{n}}+c_2\frac{G^2}{n^2L\epsilon}\sqrt{\frac{p\log(1/\delta)}{\xi}}+c_3\frac{G^2p\log(1/\delta)}{n^2\epsilon^2\mu\xi} \\
	&\quad+\left(M_\ell+\frac{G^2}{2\mu}\right)\sqrt{\frac{\log(3/\xi)}{n}},
	\end{aligned}
	\end{equation}
	for some constants $c_1,c_2,c_3$.
\end{theorem}

Detailed proof is shown in Appendix C.2.
Our proof is motivated by \cite{20}, and we extend it to differentially private pairwise learning setting.
The key challenges consist of `pairwise' and `privacy'.
As discussed before, we propose the \textit{pairwise} locally elastic stability to overcome the first problem.
And for the second problem: perturbation, we propose a \textbf{noisy version} of pairwise locally elastic stability.
See more details in Appendix C.2.

\begin{remark}
\rm{
By ignoring the constant terms in (\ref{the2}), the high probability excess population risk bound comes to:
\begin{equation*}
\mathcal{O}\left(\frac{1}{n}+\frac{1}{\sqrt{n}}+\frac{\sqrt{p}}{\sqrt{n}\epsilon}+\frac{\sqrt{p}}{n^2\epsilon}+\frac{p}{n^2\epsilon^2}\right)=\mathcal{O}\left(\frac{\sqrt{p}}{\sqrt{n}\epsilon}\right).
\end{equation*}
Our result achieves the same order as results given in \cite{4,17}.
Moreover, different from the results given in previous works, our result does not assume the loss function to be convex.
Thus, our result is more general.
}
\end{remark}

\begin{remark}
\rm{
In (\ref{the2}), on the right hand side of the inequality, the first line is the original generalization error, the second line is the excess empirical risk and the bias caused by perturbation, and the third line is the gap between $\theta^*$ and $\theta_\mathcal{P}^*$.
When comparing with the result proposed in \cite{4}, the last two terms are similar.
And to appreciate the advantages of \textbf{pairwise locally elastic stability}, we compare the generalization error term (the first line) with \cite{4}.

In \cite{4}, the generalization error is analyzed via uniform argument stability (UAS) (\cite{35}).
The UAS bounds the gap between model parameters derived from adjacent datasets: $\mathcal{A}(D)$ and $\mathcal{A}(D')$, formally defined as:
\begin{equation*}
\|\mathcal{A}(D)-\mathcal{A}(D')\|_2\leq\kappa.
\end{equation*}
If the ineuqality holds, then $\mathcal{A}$ is $\kappa$-uniformly argument stable.
Via $G$-Lipschitz property, it is easy to get that the uniform stability parameter $\beta_n^U=G\kappa$.
As a result, the generalization error in \cite{4} is:
\begin{equation*}
\frac{4\beta_n^U}{G}+48\sqrt{6}e\beta_n^U\log(n)\log(2e/\xi)+12\sqrt{2}e(M_\ell+G\|\theta_T\|_2)\sqrt{\frac{\log(2e/\xi)}{n}}.
\end{equation*}

As discussed above, $\beta_n^U=\sup_{z,z',z''}\beta_n(z;z',z'')$ and $\beta_n(z;z',z'')=\beta(z;z',z'')/n$.
So $n\beta_n^U=\sup_{z,z',z''}\beta(z;z',z'')$.

Comparing with our result:
\begin{equation*}
\frac{2\sup_{z\in\mathcal{D}}\mathbb{E}_{z',z''}\beta\left(z;z',z''\right)}{n}+\left(4\sup_{z\in\mathcal{D}}\mathbb{E}_{z',z''}\left[\beta\left(z;z',z''\right)\right]+4M_\ell\right)\sqrt{\frac{2\log\left(12/\xi\right)}{n}},
\end{equation*}
the first term $\frac{4\beta_n^U}{G}>\frac{2\sup_{z\in\mathcal{D}}\mathbb{E}_{z',z''}\beta\left(z;z',z''\right)}{n}$, which is easy to follow by the definition of $\beta_n^U$.
Under the case that the worst $\left|\ell\left(\mathcal{A}_D;z,z'\right)-\ell\left(\mathcal{A}_{D^{-i}};z,z'\right)\right|$ is much larger than the expectation ones, the gap between $n\beta_n^U$ and $\sup_{z\in\mathcal{D}}\mathbb{E}_{z',z''}\beta\left(z;z',z''\right)$ is much larger, leading our result much better.

In \cite{4}, $\beta_n^U=\sqrt{T}/n$ and $T>n$, so the second term $48\sqrt{6}e\beta_n^U\log(n)\log(2e/\xi)$ is a $\mathcal{O}\left(\log(n)/\sqrt{n}\right)$ term.
Thus, we omit it here.

Recall that $\beta_n^U=\sqrt{T}/n$.
For the third term, considering that we take $T=\mathcal{O}\left(\log(\sqrt{n}\epsilon)\right)$ in Theorem 3, we have $\sup_{z,z',z''}\beta(z;z',z'')=n\beta_n^U=\mathcal{O}(\log^{1/2}\left(\sqrt{n}\epsilon\right))$, which is a small value.
Additionally, $\sup_{z\in\mathcal{D}}\mathbb{E}_{z',z''}\beta\left(z;z',z''\right)$ is (much) less than $\sup_{z,z',z''}\beta(z;z',z'')$, and term $\|\theta_T\|_2$ always contains term $p$.
As a result, term $\left(4\sup_{z\in\mathcal{D}}\mathbb{E}_{z',z''}\left[\beta\left(z;z',z''\right)\right]+4M_\ell\right)\sqrt{\frac{2\log\left(12/\xi\right)}{n}}$ in our result is similar to or better than term $12\sqrt{2}e(M_\ell+G\|\theta_T\|_2)\sqrt{\frac{\log(2e/\xi)}{n}}$ derived from \cite{4} (unless $n$ is larger than the order of $\mathcal{O}(e^{36e^2})$).

Combining the three terms mentioned above together, our result is much better, which means that the result derived by our proposed pairwise locally elastic stability is better than which derived via pairwise uniform stability.
}
\end{remark}

\begin{table*}
	\centering
	\caption{Utility bounds of our method and previous pairwise learning methods}
	\begin{threeparttable}
		\begin{tabular}{c|ccccc|c}
			\hline
			& $G$ & $L$ & $\alpha$ & C & PL & Excess Population Risk \\
			\hline
			\cite{17} (High Probability) & $\checkmark$ & $\checkmark$ & $\checkmark$ & $\times$ & $\times$ & $\mathcal{O}\left(\frac{\sqrt{p}}{n\epsilon}\right)$ \\
			\hline
			\cite{17} (High Probability) & $\checkmark$ & $\checkmark$ & $\times$ & $\checkmark$ & $\times$ & $\mathcal{O}\left(\frac{\sqrt{p}}{\sqrt{n}\epsilon}\right)$ \\
			\hline
			\cite{4} (High Probability) & $\checkmark$ & $\times$ & $\times$ & $\checkmark$ & $\times$ & $\mathcal{O}\left(\frac{\sqrt{p}}{\sqrt{n}\epsilon}\right)$ \\
			\hline
			Our Proposed (Expectation) & $\checkmark$ & $\checkmark$ & $\times$ & $\times$ & $\checkmark$ & $\mathcal{O}\left(\frac{\sqrt{p}}{n\epsilon}\right)$ \\
			\hline
			Our Proposed (High Probability) & $\checkmark$ & $\checkmark$ & $\times$ & $\times$ & $\checkmark$ & $\mathcal{O}\left(\frac{\sqrt{p}}{\sqrt{n}\epsilon}\right)$ \\
			\hline
		\end{tabular}
		\label{tab2}
	\end{threeparttable}
\end{table*}

\section{Comparisons}

The comparisons between our method and previous methods (\cite{17,4}) are shown in Table \ref{tab2}.
In previous works, output perturbation is applied, in which random noise is added only once, directly to the model.
In this paper, we propose a gradient perturbation method, introducing a new paradigm to differentially private pairwise learning.
Because in gradient perturbation method, random noise is added to the gradient for $T$ times, so the analysis is more complicated.

For the expectation bounds, we give an excess population risk bound of $\mathcal{O}\left(\sqrt{p}/(n\epsilon)\right)$, better than previous results.
It is worth emphasizing that the expectation bounds do not assume the loss function to be convex, and we achieve similar excess population risk bound to previous result under strongly convex condition, and our result is better than previous convex results by a factor of $\mathcal{O}\left(\sqrt{n}\right)$.

For the high probability bound, we give an excess population risk bound of $\mathcal{O}\left(\sqrt{p}/(\sqrt{n}\epsilon)\right)$.
Like mentioned above, this bound can also be applied to non-convex condition.
And our result achieves previous results under convexity assumption.
Moreover, our result is better than previous results because `pairwise locally elastic stability' is applied, as discussed in Remark 4.

Additionally, as shown in Table \ref{tab2}, our proposed expectation excess population risk bound is better than the high probability bound, the gap is of the order $\mathcal{O}\left(\sqrt{n}\right)$.

\subsection*{Technological Novelty}
\paragraph{For Expectation Bounds:}
\textit{On-average stability} was proposed to analyze the generalization error in convex non-DP pairwise learning setting.
The barriers to extend it to our setting include:
1) Previous analysis assumed that the loss function is $\alpha$-strongly convex, then for model parameters on datasets $D$ and $D_i$, $\ell\left(\theta^*_i\right)-\ell\left(\theta^*\right)\geq\frac{\alpha}{2}\|\theta^*_i-\theta^*\|_2^2$ holds ((B.3) in \cite{24}), so $\mathbb{E}\left[\|\theta^*_i-\theta^*\|_2^2\right]$ can be easily bounded.
However, when it comes to non-convex condition, the inequality does not hold;
2) Previous analysis is based on gradient descent, without random noise, which is an essential part in DP.
So, in our analysis, except for different datasets, bias are also brought to the model parameters from the random noise.
To solve these problems, in this paper, we track the model parameters at each iteration, to: 1) measure the gap between $\theta_i^T$ and $\theta^T$;
2) analyze the bias brought by the random noise, and the expectation is got by Jessen inequality.

\paragraph{For High Probability Bounds:}
\textit{Locally elastic stability} is an approach to measure the difference of the loss functions on models derived from adjacent datasets (we call them \textbf{adjacent models} here).
It was proposed to improve the generalization error in non-DP pointwise learning setting.
The barriers to extend it to our setting include:
1) The locally elastic stability analysis is based on influence function.
However, the influence function is only discussed in pointwise learning, corresponding framework for pairwise learning was still an open question.
2) In the field of DP, random noise is injected to the algorithm and adjacent models are changed by noise, so it is not easy to apply locally elastic stability to privacy condition.
3) When analyzing the high probability excess empirical risk bound, there exists term $\left\Vert b\right\Vert_2^2$, which does not obey Gaussian distribution.
For the expectation bound, $\mathbb{E}\left\Vert b\right\Vert_2^2$ can be got by $\mathbb{E}\left[X^2\right]=\mathbb{E}\left[X\right]+var(X)$, but in high probability condition, this solution cannot be applied.
To solve these problems, in this paper, we:
1) construct a pairwise influence function paradigm as a basis of locally elastic stability analysis, filling the blank of pairwise influence function;
2) propose a noisy version of elastic stability, considering the worst case: when most noise is injected to adjacent models, and the high probability bound of the noise is got by Chebyshev's inequality and Markov's inequality.
3) observe that $\left\Vert b\right\Vert_2^2=\sum_{i=1}^{p}(b_{(i)})^2$, so we transform it into Gamma distribution to solve the problem.

\section{Examples}

\begin{example}[Bipartite Ranking (\cite{31})]
With $\left\Vert\theta\right\Vert_2\leq1$,
the pairwise loss function is
$\ell\left(\theta;z,z'\right)=\phi\left(\left(y-y'\right)h(\theta;x,x')\right)$, where $h(\theta;x,x')=\theta^T(x-x')$, $\mathcal{C}=\{\theta:\theta\in\mathbb{R}^p,\left\Vert\theta\right\Vert_2\leq 1\}$, and $\phi(x)=\log\left(1+e^{-x}\right)$.
\end{example}

\begin{example}[Metric Learning (\cite{30})]
	To learn a Mahalianobios metric $M_\theta^2(x,x')=\left(x-x'\right)^T\theta(x-x')$, the loss function of metric learning is $\ell(\theta;z,z')=\phi\left(yy'(1-M_\theta^2(x,x'))\right)$ with the same $\phi$ in Example 1, where $y,y'\in\{-1,+1\}$, $\mathcal{C}=\{\theta:\theta\in\mathbb{R}^{p*p},\left\Vert\theta\right\Vert_2\leq 1\}$.
\end{example}

If a regularization term $\lambda\|\theta\|_2^2$ is added to $\ell(\cdot;\cdot,\cdot)$ mentioned in Examples 1 and 2, the loss functions obviously satify PL condition with parameter $\lambda$.

\begin{lemma}
If $x$ and $y$ are bounded and one runs Algorithm 1 with $T=\mathcal{O}(\log(\sqrt{n}\epsilon))$ and $\eta=1/L$, then with high probability we have:
$L_\mathcal{P}(\theta_{priv})-L_\mathcal{P}(\theta_\mathcal{P}^*)=\mathcal{O}(\sqrt{p}/(\sqrt{n}\epsilon))$.
If $T=\mathcal{O}(\log(n\epsilon/\sqrt{p}))$, the expectation bound of $L_\mathcal{P}(\theta_{priv})-L_\mathcal{P}(\theta_\mathcal{P}^*)$ is $\mathcal{O}(\sqrt{p}/(n\epsilon))$, for Bipartite Ranking problem.
\end{lemma}

For example, if $\left\Vert x\right\Vert_2\leq1$ and $y,y'\in[-1,+1]$, then $G=2$, $L=\frac{2+2\sqrt{2}}{3+2\sqrt{2}}\approx0.84$.
So if applying Theorems 2 and 3, the lemma holds.

\begin{lemma}
	If $x$ is bounded, then the results given Lemma 2 hold for metric learning problem.
\end{lemma}
For example, if $\left\Vert x\right\Vert_2\leq1$, then $G=1$, $L=\frac{1+\sqrt{2}}{6+4\sqrt{2}}\approx0.21$.
Then Lemma 3 holds.

For the same loss function, if the examples above are applied to neural networks with one hidden layer, then they satisfy PL inequality.
And considering that neural networks are based on function composition, if the activation is differentiable and twice differentiable, then it satisfies $G$-Lipschitz and $L$-smooth.
Moreover, if $L$-smooth cannot be guaranteed, there are some smoothing techniques that can be performed, such as Moreau envelope smoothing (\cite{19}).
For more complicated models, if PL inequality cannot be satisfied, the privacy guarantees and the generalization error bound (analyzed in the Appendix) still work because they do not rely on the assumption of PL condition.

\section{Conclusions}

In this paper, we propose a gradient perturbation method for DP pairwise learning, introducing a new paradigm to the field of DP pairwise learning.
We first analyze the privacy guarantees via the moments accountant method.
Then, we analyze the excess population risk and give corresponding expectation and high probability bounds.
For the expectation and high probability bounds, we do not assume the loss function to be convex.
To the best of our knowledge, this is the first non-convex analysis over DP pairwise learning.
By detailed theoretical analysis, the utlity bounds of our method is similar to (or better than) the results proposed by the previous methods under the convex (or even strongly convex) assumption.

\bibliographystyle{plainnat}
\bibliography{neurips_2021}

\end{document}